\documentclass[10pt,twocolumn,letterpaper]{article}

\usepackage{iccv}
\usepackage{times}
\usepackage{epsfig}
\usepackage{graphicx}
\usepackage{amsmath}
\usepackage{amssymb}
\usepackage{bm}
\usepackage{color}
\usepackage{multirow}
\usepackage{diagbox}
\usepackage{cancel}

\newcommand{\1}[1]{{\bf{\color{red}#1}}}
\newcommand{\2}[1]{{\bf{\color{blue}#1}}}
\usepackage[pagebackref=true,breaklinks=true,letterpaper=true,colorlinks,bookmarks=false]{hyperref}

\iccvfinalcopy 

\def\iccvPaperID{695} 

\ificcvfinal\fi
\begin{document}
\title{Class Rectification Hard Mining for Imbalanced Deep Learning}

\author{Qi Dong\\
{\small Queen Mary University of London}\\
{\tt\small q.dong@qmul.ac.uk}
\and
Shaogang Gong\\
{\small Queen Mary University of London}\\
{\tt\small s.gong@qmul.ac.uk}
\and Xiatian Zhu\\
{\small Vision Semantics Ltd.}\\
\tt\small eddy@visionsemantics.com
}

\maketitle

\begin{abstract}
Recognising detailed facial or clothing attributes
in images of people is a challenging task for computer vision, 
especially when the training data are both in very large scale and
extremely imbalanced among different attribute classes. 
To address this problem, we formulate a novel scheme for batch
incremental hard sample mining of 
minority attribute classes from imbalanced large scale training data.
We develop an end-to-end deep learning framework capable of 
avoiding the dominant effect of majority classes
by discovering sparsely sampled boundaries of minority classes.
This is made possible by introducing a Class Rectification Loss (CRL) regularising algorithm.
We demonstrate the advantages and scalability of CRL over existing
state-of-the-art attribute recognition and imbalanced data learning
models on two large scale imbalanced benchmark datasets, the CelebA facial
attribute dataset and the X-Domain clothing attribute dataset.
	
\end{abstract}

\section{Introduction}
\label{sec:Intro}
\begin{figure} 
	\includegraphics[width=1\linewidth]{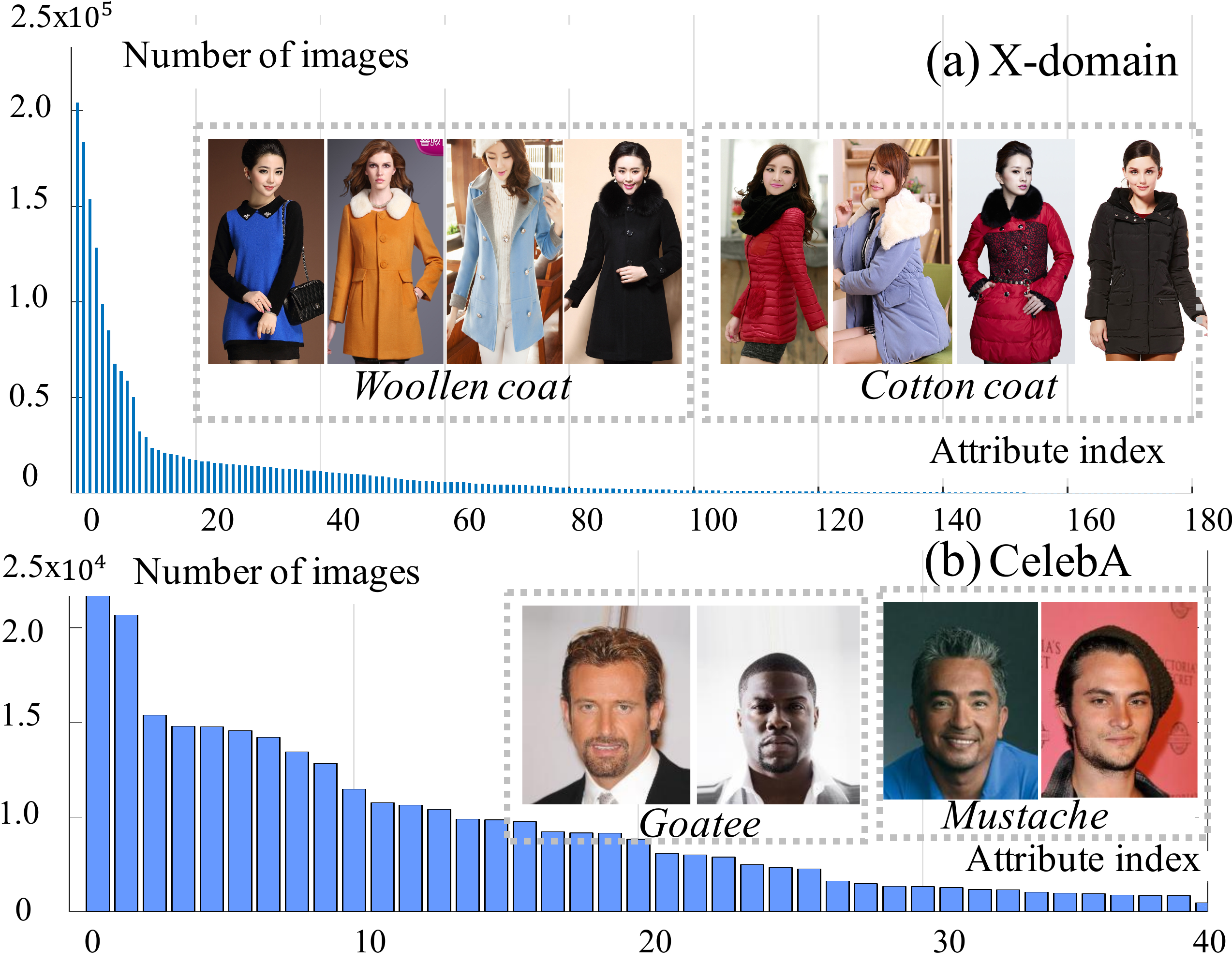}
	\caption{Imbalanced training data distribution: (a) clothing
          attributes (X-Domain \cite{chen2015deep}), (b) facial attributes 
		(CelebA \cite{liu2015deep}). 
}          
	\label{fig:problem}
	\vspace{-0.3cm}
\end{figure}

Automatic recognition of person attributes in images, e.g. clothing
category and facial characteristics, is very 
useful \cite{gong2014person,feris2014attribute}, but also
challenging due to: 
(1) Very large scale training data with significantly imbalanced
distributions on annotated attribute data
\cite{Akbani-ecml04,ChenEtAlcvpr13,huang2016learning},
with clothing and face
attributes typically exhibiting a power-law distribution (Figure
\ref{fig:problem}). 
This makes model learning biased towards 
well-labelled attribute classes (the {\em majority classes}) resulting in
poor performance against sparsely-labelled classes (the {\em minority
  classes}) \cite{he2009learning}, known as the {\em imbalanced class
  learning} problem \cite{he2009learning}. 
(2) Subtle discrepancy between different fine-grained attributes, e.g. 
``Woollen-Coat'' can appear very similar to ``Cotton-Coat'', whilst ``Mustache" may be visually indistinct (Figure \ref{fig:problem}). 
To recognise such subtle attribute differences,
model training {\em assumes} a large collection of {\em balanced} training
image data \cite{chen2015deep,zhang2014panda}.

\begin{table*} [th] 
	\footnotesize
	\centering
	\setlength{\tabcolsep}{0.08cm}
	\caption{\footnotesize
Comparing large scale datasets in terms of training data imbalance. Metric:
the size ratio of smallest and largest classes. 
These numbers are based on the standard
train data split if available, otherwise on the whole
dataset. For COCO \cite{lin2014microsoft}, no specific numbers are available
for calculating between-class imbalance ratios, mainly because the
COCO images often contain simultaneously multiple classes of objects 
and also multiple instances of a specific class. }
\label{tab:dataset_imbalance}
	\begin{tabular}{c||c|c|c|c|c||c|c|c}
		\hline
		Datasets 
		& ILSVRC2012-14 \cite{russakovsky2015imagenet} 
		%
		& COCO \cite{lin2014microsoft} 
		& VOC2012 \cite{everingham2015pascal}
		& CIFAR-100 \cite{krizhevsky2009learning}
		& Caltech 256 \cite{griffin2007caltech} 
		& CelebA \cite{liu2015deep} 
		& DeepFashion \cite{liu2016deepfashion}
		& X-Domain \cite{chen2015deep}  
		\\ \hline \hline 
		Imbalance ratio & 1 : 2 
		& - & 1 : 13 
		& 1 : 1 & 1 : 1 & 1 : 43 
		&  1 : 733 
		& \bf 1 : 4162
		\\ \hline

		\hline 
	\end{tabular}
\vspace{-0.55cm}
\end{table*}

There have been studies on how to solve the general imbalanced
data learning problem including re-sampling
\cite{chawla2002smote,maciejewski2011local,oquab2014learning} and cost-sensitive weighting
\cite{ting2000comparative,tang2009svms}. 
However, these methods can suffer from either {\em over-sampling}
which leads to model overfitting and/or introducing noise, or
{\em down-sampling} which loses valuable data.
These classical imbalanced learning models rely typically on
hand-crafted features, without deep learning's capacity
for exploiting a very large pool of imagery data from diverse sources
to learn more expressive representations
\cite{simonyan2014very,sharif2014cnn,krizhevsky2012imagenet,bengio2013representation}.
However, deep learning is likely to suffer even more from imbalanced
data distribution 
\cite{zhou2006training,jeatrakul2010classification,khan2015cost,huang2016learning}
and deep learning of imbalanced data is currently under-studied. 
This is partly due to that popular image datasets for deep learning,
e.g. ILSVRC,  
do not exhibit significant class imbalance due to careful data 
filtering and selection during 
the construction process (Table \ref{tab:dataset_imbalance}). The
problem becomes very challenging for deep learning of clothing or facial
attributes (Figure~\ref{fig:problem}).
In particular, when a
large scale training data are drawn from 
online Internet sources
\cite{chen2015deep,huang2015cross,liu2016deepfashion,liu2015deep},
image attribute distributions are likely to be extremely imbalanced
(see Table \ref{tab:dataset_imbalance}).
For example, the data sampling size ratio between the minority and
majority classes (imbalance ratio) in the X-Domain clothing attribute
dataset \cite{chen2015deep} is 1:4,162, 
with the smallest minority and largest majority class
having 24 and 99, 885 
images respectively.

This work addresses the problem of deep learning on large scale imbalanced person
attribute data for multi-label attribute recognition. 
Other deep models for imbalanced data learning exist
\cite{zhou2006training,jeatrakul2010classification,oquab2014learning,khan2015cost}. 
These models shall be considered as end-to-end deep feature learning  and classifier learning. For over-sampling and down-sampling, a special training data re-sampling pre-process may be needed prior to deep model learning.
They are ineffective for deep learning of imbalanced data
(see evaluations in Sec.~\ref{sec:exp}).
More recently, a Large Margin Local Embedding (LMLE) method
\cite{huang2016learning} was proposed to enforce the local
cluster structure of per class distribution in
the deep learning process so that minority classes can better maintain
their own structures in the feature space.
The LMLE has a number of fundamental drawbacks including 
disjoint feature and classification optimisation, offline
clustering of training data {\em a priori} to model learning,
and quintuplet construction updates. 


This work presents a novel {\em end-to-end} deep learning approach to modelling
multi-label person attributes, clothing or facial, given a large scale
webly-collected image data pool with significantly imbalanced
attribute data distributions. The {\bf contributions} of this work are:
(1) We propose a novel model for deep learning of very large scale imbalanced data based on
{\em batch-wise incremental hard mining} of hard-positives and hard-negatives from
minority attribute classes alone.
This is in contrast to existing attribute recognition methods \cite{chen2015deep,huang2015cross,liu2016deepfashion,dong2016multi,zhang2014panda}
which either assume balanced training data or simply ignore the
problem. Our model performs an end-to-end feature
representation and multi-label attribute classification joint learning. 
(2) We formulate a {\em Class Rectification Loss} (CRL)
regularising algorithm. This is designed to explore the per batch sampled
hard-positives and hard-negatives for improving minority class
learning with batch-balance updated deep features. 
Crucially, this loss rectification is correlated explicitly with
batch-wise (small data pool) iterative 
model optimisation therefore achieving incremental imbalanced data learning
for all attribute classes. 
This is in contrast to LMLE's global clustering of the entire training data
(large data pool) and ad-hoc estimation of cluster size. 
Moreover, given our batch-balancing hard-mining approach, the proposed CRL is
independent to the overall training data size, therefore very scalable to
large scale training data.
Our extensive experiments on two large scale datasets CelebA
\cite{liu2015deep} and X-Domain \cite{chen2015deep}
against 11 different models including 7 state-of-the-art deep
attribute models demonstrate the advantages of the proposed method.

\noindent {\bf Related Work. }
{\em Imbalanced Data Learning.} There are two classic approaches to
learning from imbalanced data, (1) {\em 
	Class re-sampling}: Either
down-sampling the majority class or over-sampling the minority class
or both \cite{chawla2002smote, drummond2003c4, han2005borderline,
	he2009learning, maciejewski2011local, oquab2014learning}. 
However, over-sampling can easily introduce undesirable noise and also
risk from overfitting. Down-sampling is thus often preferred, but this may
suffer from losing valuable information
\cite{drummond2003c4}. 
(2) {\em Cost-sensitive learning}: Assigning higher misclassification
costs to the minority classes as compared to the majority classes
\cite{ting2000comparative, zadrozny2003cost, chen2004using,
	zhou2006training, tang2009svms}, or regularising the cross-entropy loss 
to cope with the imbalanced positive and negative class distribution \cite{shen2015deepcontour}.
For this kind of data biased learning, most commonly adopted in deep
models is positive data augmentation, e.g. to learn a deep
representation embedding the local feature structures of minority
labels \cite{huang2016learning}.
{\em Hard Mining.} 
Negative mining has been used for pedestrian detection \cite{felzenszwalb2010object},
face recognition \cite{schroff2015facenet},
image categorisation \cite{oh2016deep,wang2014learning,cui2016fine},
unsupervised visual representation learning \cite{wang2015unsupervised}. 
Instead of general negative mining, the rational for mining {\em hard}
negatives (unexpected) is that they are more informative than {\em
  easy} negatives (expected). Hard negative
mining enables the model to improve itself quicker and more
effectively with less data. Similarly, model learning can also benefit
from mining hard positives (unexpected). In our model learning we {\em
  only} consider hard mining on the minority classes for efficiency
therefore our batch-balancing hard mining strategy differs
significantly from that of LMLE \cite{huang2016learning} in 
that: (1) The LMLE requires to exhaustively search the entire training
set and thus less scalable to large sized data due to computational cost;
(2) Hard mining in LMLE is on {\em all}
classes, both the minority and the majority classes, therefore not
strictly focused on imbalanced learning of the minority classes
thus more expensive whilst less effective. 
{\em Deep Learning of Person Attributes.}
Personal clothing and/or facial attributes are key to person description.
Deep learning have been exploited for clothing 
\cite{chen2015deep,huang2015cross,liu2016deepfashion,dong2016multi,wang2017Attribute}
and facial attribute recognition \cite{liu2015deep,zhang2014panda} 
due to the availability of large scale datasets and deep models'
capacity for learning from large sized data. 
However, these methods mostly ignore the significantly imbalanced class
data distributions, resulting in suboptimal model
learning for the minority classes. 
One exception is the LMLE model \cite{huang2016learning} which explicitly
considers the imbalanced attribute class learning challenge.
In contrast to our end-to-end deep learning model in this work,
LMLE is not end-to-end learning and suffers from poor scalability and
suboptimal optimisation. This is due to LMLE's need for very expensive
quintuplet construction and pre-clustering (suboptimal) on the entire training data,
resulting in separated feature and classifier learning.

\section{Class Rectification Deep Learning}
\label{sec:method}


\begin{figure}[th]
	\centering
	\includegraphics[width=1\linewidth]{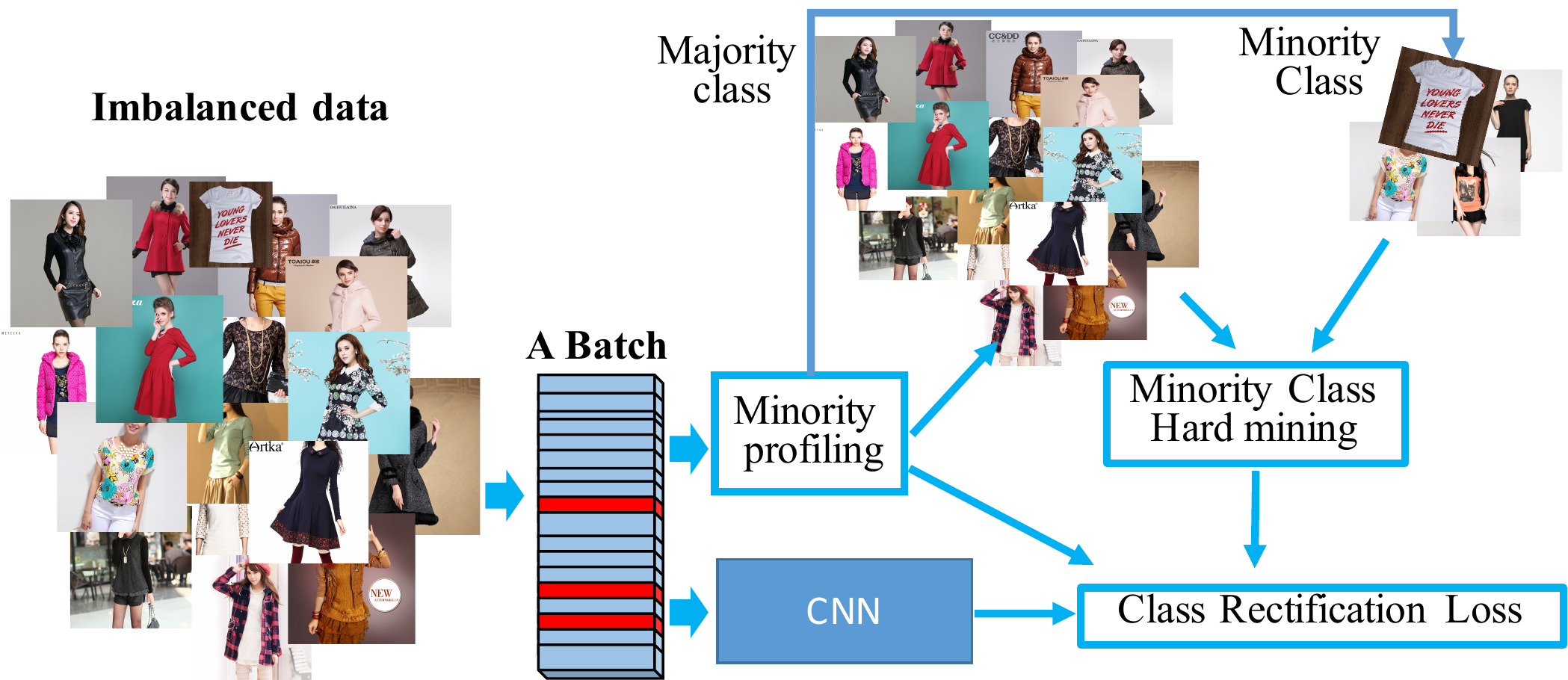}
	\caption{Overview of our Class Rectification Loss (CRL) regularising
		approach for deep end-to-end imbalanced data learning. 
}
	\vspace{-0.15cm}
	\label{fig:pipeline}
\end{figure}

We wish to construct a deep model capable of recognising multi-labelled person attributes
$\{z_j\}_{j=1}^{n_\text{attr}}$ in
images, with a total of $n_\text{attr}$ different attribute categories, 
each category $z_j$ having its respective value range $Z_j$,
e.g. multi-valued (1-in-N) clothing category or binary-valued (1-in-2)
facial attribute. 
%
%
Suppose that we have a collection of $n$ training images
$\{\bm{I}_i\}_{i=1}^{n}$ along with their attribute annotation vectors
$\{\bm{a}_i\}_{i=1}^{n}$,
and $\bm{a}_i=[a_{i,1}, \dots, a_{i,j}, \dots,a_{i,n_\text{attr}}]$
where $a_{i,j}$ refers to the $j$-th attribute value of the image $\bm{I}_i$.
The number of image samples available for different attribute classes
varies greatly (Figure~\ref{fig:problem}) therefore poses a
significant {\em imbalanced data} distribution challenge to model learning.
Most attributes are {\em localised} to image regions, even though the
location information is not provided in the annotation ({\em weakly labelled}).
Intrinsically, this is a {\em multi-label} recognition problem since
the $n_\text{attr}$ attributes may co-exist in every person
image. 
%
To that end, 
we consider to jointly learn {\em end-to-end} features and {\em all}
the attribute classifiers given imbalanced image data. 
Our method can be readily incorporated with the classification loss function (e.g. Cross-entropy loss) 
of standard CNNs without the need for a new optimisation algorithm (Fig. \ref{fig:pipeline}).

\noindent {\em Cross-entropy Classification Loss.}
For multi-class classification CNN model training 
(CNN model details in ``Network Architecture'',
Sec.~\ref{sec:eval_face_attributes} and \ref{sec:eval_clothing_attr}),
one typically considers the Cross-entropy loss function 
by firstly predicting the $j$-th attribute posterior probability of image $\bm{I}_i$
over the ground truth $a_{i,j}$:
\begin{equation}
{p}(y_{i,j} = a_{i,j} | \bm{x}_{i,j}) = \frac{\exp(\bm{W}_{j}^{\top} \bm{x}_{i,j})} {\sum_{k=1}^{|Z_{j}|} \exp(\bm{W}_{k}^{\top} \bm{x}_{i,j})}
\end{equation}
where $\bm{x}_{i,j}$ refers to the feature vector of $\bm{I}_i$ for $j$-th attribute,
and $\bm{W}_k$ is the corresponding prediction function parameter. We then compute
the overall loss on a batch of $n_\text{bs}$ images as
the average additive summation of attribute-level loss with equal weight:
\begin{equation}
l_\text{ce} = - \frac{1}{n_\text{bs}}\sum_{i=1}^{n_\text{bs}}  \sum_{j=1}^{n_\text{attr}} \log \Big(p(y_{i,j}=a_{i,j}|\bm{x}_{i,j}) \Big)
\label{eq:loss}
\end{equation}
However, given highly imbalanced image samples on different attribute classes,
model learning by the conventional classification loss is suboptimal. 
To address this problem, we reformulate the model 
learning objective loss function by mining explicitly in each batch of
training data both hard positive
and hard negative samples for every minority attribute class. Our
objective is to rectify incrementally per batch the class bias in model learning so
that the features are less biased towards the 
over-sampled majority classes and more sensitive to the class
boundaries of under-sampled minority classes.

\subsection{Minority Class Hard Mining}
\label{sec:method_hard_mining}

We wish to impose minority-class hard-samples as constraints on the
model learning objective. 
Different from the approach adopted by the LMLE model
\cite{huang2016learning} which aims to 
preserve the local structures of {\em both} majority and minority classes by global
sampling of the entire training dataset, we explore {\em batch-based} hard-positive
and hard-negative mining for the minority classes {\em only}. We do not assume the local
structures of minority classes can be estimated from global clustering
before model learning. To that end, we consider the following steps
for handling data imbalance.

\noindent {\bf Batch Profiling of Minority and Majority Classes. } 
In each training batch, we profile to discover the minority and majority classes.
Given a batch of $n_\text{bs}$ training samples,
we profile the attribute class distribution $\bm{h}^j = [h_1^j, \dots, h_k^j, \dots h_{|Z_i|}^j]$ 
over $Z_j$ for each attribute $j$,
where $h_k^j$ denotes the number of training samples with the $j$-th attribute class value assigned to $k$.
Then, we sort $h_k^j$ in the descent order.
As such, we define minority classes {\em in this batch} as those classes $C_\text{min}^i$ with
the smallest number of training samples, with the condition that
\begin{equation}
\sum_{k \in C_\text{min}^j} h_k^j < 0.5n_\text{bs}.
\end{equation} 
That is, all minority classes only contribute to less than half of the
total data samples in this batch.
The remaining classes are deemed as the majority classes.

We then exploit a minority class hard mining scheme to facilitate
additional loss constraints in model learning\footnote{
	We consider only those minority classes having at least two
	sample images in each batch, ignoring those minority classes having only one sample
	image or none. This enables triplet loss based learning.}.
\begin{figure}[th]
	\centering
	\includegraphics[width=0.45\textwidth]{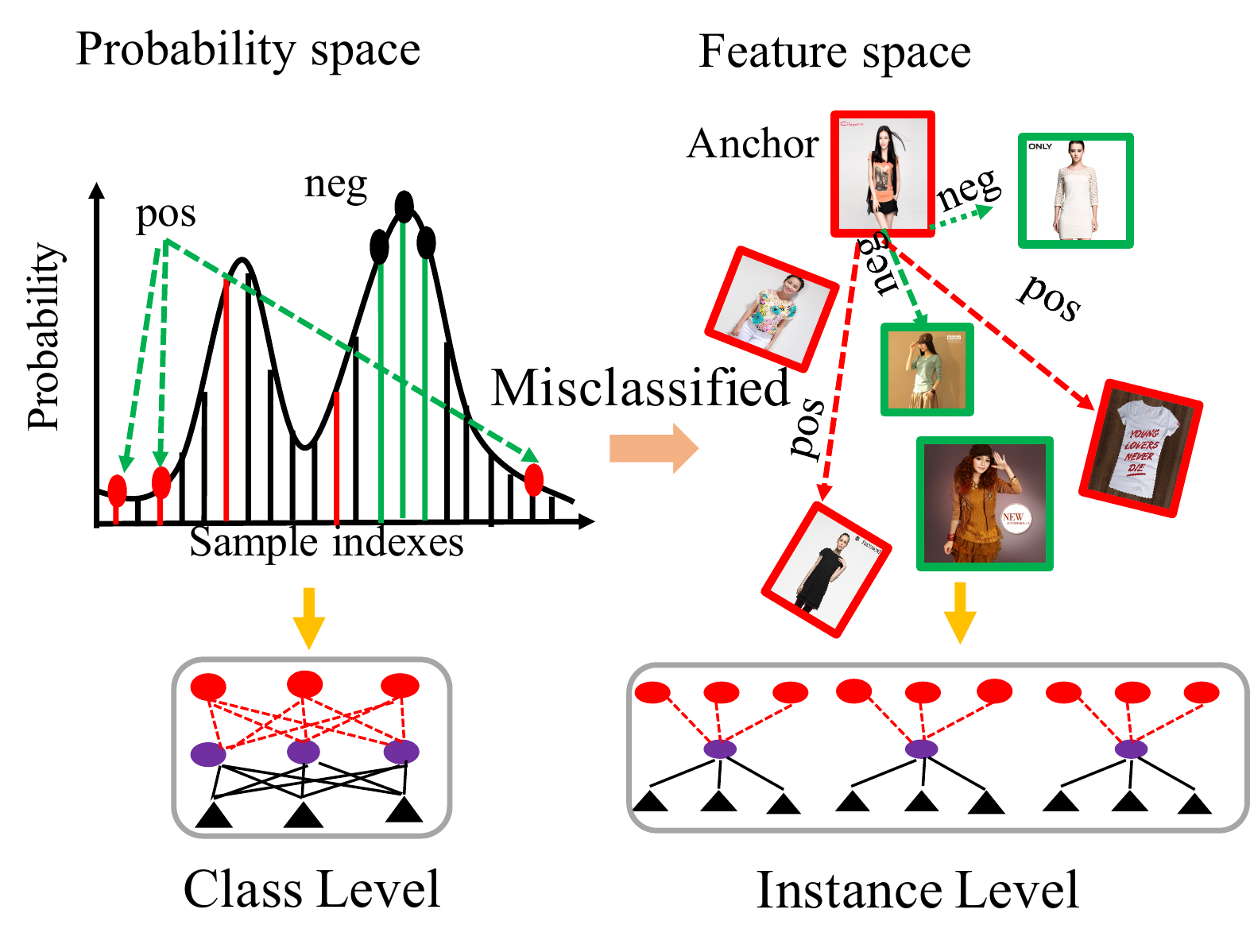}
	\vskip -0.2cm
	\caption{Illustration of the proposed minority class hard mining.
	}	
	\vspace{-0.4cm}
	\label{fig:hardmining}
\end{figure}
To that end, we consider two approaches:
(I) Minority class-level hard mining (Fig. \ref{fig:hardmining}(left)), 
(II) minority instance-level hard mining (Fig. \ref{fig:hardmining}(right)).

\noindent {\bf (I) Minority Class-Level Hard Samples. } 
At the class level, 
for a specific minority class $c$ of attribute $j$, 
we refer ``hard-positives'' to those images $\bm{x}_{i,j}$ from class $c$
($a_{i,j} = c$ with $a_{i,j}$ denoting the attribute $j$ ground truth label of $\bm{x}_{i,j}$)
given {\em low} discriminative scores 
$p(y_{i,j}=c|\bm{x}_{i,j})$
on class $c$ 
by the model,
i.e. {\em poor} recognitions. 
Conversely, by ``hard-negatives'', we refer to those images $\bm{x}_{i,j}$ from 
other classes ($a_{i,j} \neq c$)
given {\em high} discriminative scores 
on class $c$ 
by the model,
i.e. {\em obvious} mistakes. 
Formally, we define them as:
\begin{equation}
\mathcal{P}^\text{cls}_{c,j} = \{ \bm{x}_{i,j} | 
a_{i,j}=c, \; 
\text{low } p(y_{i,j}=c|\bm{x}_{i,j})
 \}
\end{equation} 
\begin{equation}
\mathcal{N}^\text{cls}_{c,j} = \{ \bm{x}_{i,j} | 
a_{i,j} \neq c, \;  
\text{high } p(y_{i,j}=c|\bm{x}_{i,j})  
 \}
\end{equation}
where 
$\mathcal{P}^\text{cls}_{c,j}$ and $\mathcal{N}^\text{cls}_{c,j}$ denote the hard positive and negative sample sets of a minority class $c$ of attribute $j$.
%

\noindent {\bf (II) Minority Instance-Level Hard Samples.}
At the instance level, we consider hard positives and negatives for each specific 
sample instance $\bm{x}_{i,j}$ 
from a minority class $c$ of attribute $j$,
i.e. with $a_{i,j} = c$.
We define ``hard-positives'' of $\bm{x}_{i,j}$ as those class $c$ images
$\bm{x}_{k,j}$ (${a}_{k,j} = c$) misclassified ($\hat{a}_{k,j} \neq c$
with $\hat{a}_{k,j}$ denoting the attribute $j$ predicted label of
$\bm{x}_{k,j}$) 
by the current model with {\em large} distances (low
matching scores) from $\bm{x}_{i,j}$ in the feature space.
``Hard-negatives'' are those images $\bm{x}_{k,j}$
not from class $c$ 
($a_{k,j} \neq c$) 
with {\em small} distances (high matching scores) to $\bm{x}_{i,j}$ in
the feature space. We define them as: 
%
\begin{equation}\small
\mathcal{P}_{i,c,j}^\text{ins} = \{ \bm{x}_{k,j} | 
a_{k,j}=c, \; 
\hat{a}_{k,j} \neq c, \;
\text{large } \mbox{dist}(\bm{x}_{i,j},\bm{x}_{k,j})
\}
\end{equation} 
\begin{equation}\small
\mathcal{N}_{i,c,j}^\text{ins} = \{ \bm{x}_{k,j} |
a_{k,j} \neq c, \;
\text{small } \mbox{dist}(\bm{x}_{i,j},\bm{x}_{k,j})
\}
\end{equation}
where 
$\mathcal{P}_{i,c,j}^\text{ins}$ and $\mathcal{N}_{i,c,j}^\text{ins}$ are the hard positive and
negative sample sets of a minority class $c$ instance $\bm{x}_{i,j}$ in 
attribute $j$,
and $\mbox{dist}(\cdot)$ is the $L_2$ distance metric.

\noindent {\bf Hard Mining.}
Intuitively, mining hard-positives enables the model to discover and expand sparsely
sampled minority class boundaries, whilst mining hard-negatives aims
to improve the margins of minority class boundaries corrupted
by visually very similar imposter classes, e.g. significantly overlapped outliers.
To facilitate and simplify model training, we adopt the following mining strategy.
At training time, 
for a minority class $c$ of attribute $j$
(or a minority class instance $\bm{x}_{i,j}$) in
each training batch data,
we select $K$ hard-positives as the bottom-$K$ scored on $c$ (or
bottom-$K$ (largest) distances to $\bm{x}_{i,j}$),
and $K$ hard-negatives as the top-$K$ scored on $c$ (or top-$K$
  (smallest) distance to $\bm{x}_{i,j}$),
given the current feature space and classification model.
This hard mining strategy allows our model optimisation to concentrate particularly on 
either poor recognitions or obvious mistakes. 
This not only reduces the model optimisation complexity
by soliciting fewer learning constraints, but also minimises computing cost. 
It may seem that some discriminative information is lost by doing so.
However, it should be noted that we perform hard-mining 
{\em independently} in each batch and {\em incrementally} over successive batches.
Therefore, such seemingly-ignored information are considered over the learning iterations.
Importantly, this proposed batch-wise hard-mining avoids the global sampling on the entire training
data as required by LMLE \cite{huang2016learning} which can suffer from
both negative model learning due to inconsistency
between up-to-date deep features and out-of-date cluster boundary
structures, and high computational cost in quintuplet updating. 
In contrast, our model can be learned directly by conventional
batch-based classification optimisation algorithms using stochastic gradient descent, with no need for complex modification required
by the quintuplet based loss in the LMLE model \cite{huang2016learning}.

\subsection{Class Rectification Loss}
\label{sec:method_CRL_function}
In deep feature representation model learning, 
the key is to discover latent boundaries for individual classes
and surrounding margins between different classes in the feature space.
%
To this end, we introduce a Class Rectification Loss (CRL) regularisation
$l_\text{crl}$ to rectify the learning bias from the conventional
Cross-entropy classification loss function (Eqn.~\eqref{eq:loss}) given
class-imbalanced attribute data:
\begin{equation}
l_\text{bln}\ =\ 
l_\text{crl} + l_\text{ce}
\label{eq:final_loss}
\end{equation}
where $l_\text{crl}$ is computed from the hard positive and negative samples of the minority classes.
	We further explore three different options to formulate $l_\text{crl}$.

\noindent {\bf (I) Class Rectification by Relative Comparison.}
Firstly, we exploit the general learning-to-rank idea \cite{liu2009learning}, 
and in particular the triplet based loss.
%
Considering the small number of training samples in minority classes,
it is sensible to make full use of them in order to effectively handle the 
underlying model learning bias.
Therefore, we regard each image of these minority classes 
as an ``anchor'' to quantitatively compute the batch balancing loss regularisation. 
Specifically, for each anchor ($\bm{x}_{a,j}$), we first construct
a set of triplets based on the mined top-$K$ hard-positives and 
hard-negatives associated with the corresponding attribute class $c$
of attribute $j$, i.e. class-level hard miming, or the sample instance
itself $\bm{x}_{a,j}$, i.e. instance-level hard mining.
In this way, we form at most $K^2$ triplets $T = \{(\bm{x}_{a,j}, \bm{x}_{p,j}, \bm{x}_{n,j})_k\}_{k=1}^{K^2}$ 
w.r.t. $\bm{x}_{a,j}$, 
and a total of at most $|X_\text{min}| \times n_\text{attr} \times K^2$
triplets $T$ for all the anchors $X_\text{min}^i$ across all the minority classes.
We formulate the following triplet ranking loss function to impose
a class balancing constraint in model learning:
\begin{equation} \small
l_\text{crl} = \frac{\sum_{T}
	 \max \left( 0, \,\, m_j +
	\mbox{dist}(\bm{x}_{a,j},\bm{x}_{p,j}) - \mbox{dist}(\bm{x}_{a,j},
	\bm{x}_{n,j}) \right) }
{|T|}
%
%
\label{eq:bt}
\end{equation}
where $m_j$ denotes the class margin of attribute $j$ in feature space, 
$\mbox{dist}(\cdot)$ is the $L_2$ distance.
%
We set the class margin for each attribute $i$ as
\begin{equation}
m_j = \frac{2 \pi}{|Z_j| }
\label{eqn:margin}
\end{equation}
with $|Z_j|$ the number of all possible values for attribute $j$.

\noindent {\bf (II) Class Rectification by Absolute Comparison.}
Secondly, we consider to enforce absolute distance constraints on positive and negative
pairs of the minority classes, inspired by the contrastive loss \cite{chopra2005learning}.
Specifically, for each instance $\bm{x}_{i,j}$ in a minority class $c$
of attribute $j$, we use the mined hard sets to build 
positive $P^{+} = \{\bm{x}_{i,j}, \bm{x}_{p,j}\}$ and negative $ P^{-} = \{\bm{x}_{i,j}, \bm{x}_{n,j}\}$ pairs
in each training batch. 
Intuitively, we require the positive pairs to be at close distances 
whist the negative pairs to be far away. Thus, we define the CRL regularisation as
\begin{equation}
%
%
\begin{split}
l_\text{crl} = 0.5*
\Big(\frac{\sum_{P^{+}} \mbox{dist}(\bm{x}_{i,j}, \bm{x}_{p,j})^{2}}{|P^{+}|} + \\
\frac{\sum_{P^{-}}\max \big(m_\text{apc} - \text{dist}(\bm{x}_{i,j}, \bm{x}_{n,j}), 0 \big)^{2}}{|P^{-}|}\Big)
\end{split}
\end{equation}
where $m_\text{apc}$ controls the between-class margin ($m_\text{apc}\!=\!1$ in our experiments).
This constraint aims to optimise the boundary of the minority classes by
incremental separation from the overlapping (confusing) majority class
instances by per batch iterative optimisation.

%
%
%
%
%

\noindent {\bf (III) Class Rectification by Distribution Comparison.}
Thirdly, we formulate class rectification on the minority class
instances by modelling the {\em distribution} of positive and negative pairs
constructed as in the case of ``Absolute Comparison''
described above.
In the spirit of \cite{yosinski2014transferable}, we represent the distribution of positive $P^{+}$
and negative $P^{-}$ pair sets with histograms $H^{+} = [h^{+}_1,\cdots,h^{+}_{\tau}]$ and 
$H^{-} = [h^{-}_1,\cdots,h^{-}_{\tau}]$ of $\tau$ uniformly spaced bins $[b_1,\cdots,b_{\tau}]$.
We compute the positive histogram $H^{+}$ as
\begin{equation}
h^{+}_t = \frac{1}{|P^{+}|} \sum_{(i,j) \in P^{+}} \varsigma_{i,j,t}
\end{equation}
where
\begin{equation} \small
	\varsigma_{i,j,t} = \begin{cases}
	\frac{\text{dist}(\bm{x}_{i,j}, \bm{x}_{p,j}) - b_{t-1}}{\Delta}, \; \text{if} \; \mbox{dist}(\bm{x}_{i,j}, \bm{x}_{p,j})  \in [b_{t-1},b_t] \\
	\frac{b_{t+1} - \text{dist}(\bm{x}_{i,j}, \bm{x}_{p,j})}{\Delta}, \; \text{if} \; \mbox{dist}(\bm{x}_{i,j}, \bm{x}_{p,j})  \in [b_{t},b_{t+1}] \\ 
	0. \quad\quad\quad\quad\quad\quad\quad\;\;  \text{otherwise}
	\end{cases}
\end{equation}
and $\Delta$ defines the step between two adjacent bins.
Similarly, the negative histogram $H^{-}$ can also be computed.
To enable the minority classes distinguishable from the overwhelming majority classes,
we enforce the two histogram distributions as disjoint as possible.
We then define the CRL regularisation loss by how much overlapping between
these two histogram distributions:
\begin{equation}
l_\text{crl} = \sum_{t=1}^{\tau} \big( h^{+}_t \sum_{k=1}^{t} h_k^{-} \big)
\end{equation}
Statistically, this CRL histogram loss measures the probability that 
the distance of a random negative pair is smaller than that of a random
positive pair.
This distribution based CRL aims to optimise a model towards
mining the minority class boundary areas in a non-deterministic manner.
In our evaluation (Sec. \ref{sec:eval_CRL_loss}),
we compared the effect of these three different CRL considerations. By
default, we deploy the Relative Comparison formulation in our experiments. 

\noindent {\bf Remarks.}	 
Due to the batch-wise design, the balancing effect by our proposed regularisor
is propagated through the whole training time in an incremental manner.
The CRL approach shares a similar principle to
Batch Normalisation \cite{ioffe2015batch} for easing network optimisation.
%
%
%
%
In hard mining, we do not consider anchor points from the majority classes as
in the case of LMLE \cite{huang2016learning}. Instead, our method employs a classification
loss to learn features for discriminating the majority classes
based on that the majority classes are well-sampled for learning class
discrimination. 
Focusing the CRL {\em only} on the minority classes makes our model
computationally more efficient. 
%
%
%
%
Moreover, the computational complexity for constructing quintuplets for LMLE and
updating class clustering globally is 
$n_\text{attr} \times (k \times O(n) \times 2^{\Omega(\sqrt{n})}) +
O(n^2)$ where $\Omega$ is the lower bound complexity and $O$ the
upper bound complexity, that is, super-polynomially
proportionate to the overall training data size $n$, 
e.g. over $150,000$ in our attribute recognition problem.
In contrast, CRL loss is linear to the batch size,
typically in $10^2$, independent to the overall training size (also see
``Model Training Time'' in the experiments).

\section{Experiments}
\label{sec:exp}

\begin{table*} [h!] 
	\scriptsize
	\centering
	\setlength{\tabcolsep}{0.2cm}
	\caption{
		{Facial attributes recognition on the CelebA dataset \cite{liu2015deep}. 
		*: Imbalanced data learning models. 
		Metric: Class-balanced accuracy, i.e. mean sensitivity (\%).
	CRL(C/I): CRL with Class/Instance level hard mining. 
	The $1^\text{st}$/$2^\text{nd}$
        best results are highlighted in red/blue.}
	}
	\label{tab:arts_face}
	\begin{tabular}{c||c|c|c|c|c|c|c|c|c|c|c|c|c|c|c|c|c|c|c|c||c}
		\hline
        \backslashbox{\bf Methods}{{\bf Attributes}}&  \rotatebox{90}{Attractive} & \rotatebox{90}{Mouth Open}  & \rotatebox{90}{Smiling}  & \rotatebox{90}{Wear Lipstick}  &   \rotatebox{90}{High Cheekbones} & \rotatebox{90}{Male}  &  \rotatebox{90}{Heavy Makeup} &  \rotatebox{90}{Wavy Hair} & \rotatebox{90}{Oval Face}  &
\rotatebox{90}{Pointy Nose}  & \rotatebox{90}{Arched Eyebrows}  & \rotatebox{90}{Black Hair}  &   \rotatebox{90}{Big Lips} & \rotatebox{90}{Big Nose}  &  \rotatebox{90}{Young} &  \rotatebox{90}{Straight Hair} & \rotatebox{90}{Brown Hair} & \rotatebox{90}{Bags Under Eyes} & \rotatebox{90}{Wear Earrings} & \rotatebox{90}{No Beard} & \rotatebox{90}{}   \\ \hline
\bf Imbalance ratio (1:x)&1&   1&   1&   1& 1&   1&   2&   2& 3&   3&   3&   3&3&   3&   4&   4& 4&   4 &   4&   5\\ \hline

Triplet-$k$NN \cite{schroff2015facenet} 
& 83 & 92 & 92 & 91& 86 & 91 & 88 & 77 & 61 & 61 & 73 & 82 & 55 & 68 & 75 & 63 & 76 & 63 & 69 & 82 &  \\
 
PANDA \cite{zhang2014panda} 
& 85 & 93 & \2{98} & \2{97} & 89 & \1{99} & \2{95} & 78 & 66 & 67 & 77 & 84 & 56 & 72 & 78 & 66 & 85 & 67 & 77 & 87 &  \\
 
ANet \cite{liu2015deep}
& \2{87} & \1{96} & 97 & 95 & \2{89} & \1{99} & 96 & \2{81} & \2{67} & 69 & 76 & \2{90} & 57 & \2{78} & 84 & 69 &  \1{83} & 70 & \1{83} & 93 &  \\

DeepID2 \cite{sun2014deep}
& 78 & 89 & 89 & 92  & 84  & 94  & 88 & 73 & 63 & 66 
 & 77 & 83 &  \2{62} & 73 & 76 & 65 & 79  & 74 & 75  & 88 &  \\ \hline 
 
Over-Sampling* \cite{jeatrakul2010classification}
&77 &89 &90 &92 &84 &95 &87 &70 &63 &67 
&\2{79} &84 &61 &73 &75 &66 &82 &73 & 76&88 &  \\ 
 
Down-Sampling* \cite{mani2003knn}
&78 &87 &90 &91&80&90&89  &70  &58  &63
&70 &80 &61 &76&80&61&76  &71  &70  &88&\\ 
 
Cost-Sensitive* \cite{he2009learning}
&78 &89 &90 &91&85&93&89  &75  &64 &65
&78&85&61&74&75&67&84&74&76&88& \\
LMLE* \cite{huang2016learning} 
& \1{88} & \1{96} & \1{99} &  \1{99} & \1{92} &  \1{99} &  \1{98} & \1{83} &  \1{68} & 72 &\2{79} & \1{92} & 60 & \1{80} & \1{87} & \2{73} & \1{87} & 73 & \1{83} & \1{96} &\\ 
{\bf CRL(C)* }  &  80 &  92 &  90 &  93 &  85 &  \2{96} &  88  &  \2{81} &  \1{68} & \1{77} & \1{80}  &   88 &  \1{68} &  77 &  \2{85} &  \1{76} &  82 & \2{79} & \2{82} &  91 &  \\
{\bf CRL(I)* }  &83 &\2{95} &  93 &  94 &  89 &  \2{96} &  84  &  79  &  66 &  \2{73} &  \1{80} &  \2{90} &  \1{68} &\1{80} &  84 &  73 &  \2{86} & \2{80} & \2{83} &  \2{94}  &  \\
\hline \hline
\backslashbox{\bf Methods}{{\bf Attributes}} & \rotatebox{90}{Bangs}& \rotatebox{90}{Blond Hair} & \rotatebox{90}{Bushy Eyebrows} & \rotatebox{90}{Wear Necklace} &  \rotatebox{90}{Narrow Eyes} & \rotatebox{90}{5 o’clock Shadow} &  \rotatebox{90}{Receding Hairline} &  \rotatebox{90}{Wear Necktie} &  \rotatebox{90}{Eyeglasses} &  \rotatebox{90}{Rosy Cheeks} & \rotatebox{90}{Goatee} &  \rotatebox{90}{Chubby} &  \rotatebox{90}{Sideburns} &  \rotatebox{90}{Blurry} &  \rotatebox{90}{Wear Hat} & 
  \rotatebox{90}{Double Chin} &  \rotatebox{90}{Pale Skin} &
  \rotatebox{90}{Gray
    Hair}&\rotatebox{90}{Mustache}&\rotatebox{90}{Bald}& \bf
  \rotatebox{90}{Average} \\ \hline
  \bf Imbalance ratio (1:x)&6&   6&6&   7&8&   8&  11&  13&14&  14&  15&  16&17&  18&  19&  20&22&  23&  24&  43\\\hline

Triplet-$k$NN \cite{schroff2015facenet} 
&81& 81 & 68 & 50 & 47 & 66 & 60 & 73 & 82 & 64 & 73 & 64 & 71 & 43 & 84 & 60 & 63 & 72 & 57 & 75 &  \bf{72} \\
 
PANDA \cite{zhang2014panda} 
&92& 91 & 74 & 51 & 51 & 76 & 67 & 85 & 88 & 68 & 84 & 65 & 81 & 50 & 90 & 64 & 69 & 79 & 63 & 74 &   \bf{77} \\
 
ANet \cite{liu2015deep}
&90& 90 & \2{82} & 59 & 57 & 81 & 70 & 79 & 95 & 76 & 86 & 70 & 79 & 56 & 90 & 68 & 77 & 85 & 61 & 73 &   \bf{80} \\

DeepID2 \cite{sun2014deep}
&91& 90 & 78 & 70 & 64  & 85 & 81  & 83 & 92  & 86 & 90  
& \2{81}  & 89 & 74  & 90 & 83 & 81 & 90 & 88 & 93  &  \bf{81}  \\ \hline 
Over-Sampling* \cite{jeatrakul2010classification}&90&90 &80 &71 &65 &85 &82 &79  &91  &\2{90} &89&83 &90 &76&89&84&82&90&90&92& \bf{82}  \\ 
 
Down-Sampling* \cite{mani2003knn}
&88&85 &75 &66 &61&82&79&80  &85  & 82 &85
&78 &80 &68&90&80&78&88&60&79& \bf{78}   \\  
 
Cost-Sensitive* \cite{he2009learning}
&90&89 &79 &71 &65&84&81&82  &91  &\1{92}&86 
&82&90&76&90&84&80&90&88&93& \bf{82} \\\
LMLE* \cite{huang2016learning} 
& \1{98}& \1{99} &\2{82} &  59 &  59 & 82 & 76 & \1{90} &\1{98} &  78 &\2{95} & 79 & 88 & 59 & \1{99} & 74 & 80 & 91 & 73 &  90 &   \bf{84} \\ 
 
{\bf CRL(C)* }  
&93&   91 &   \2{82} &  \1{76} & \2{70} & \2{89} & \2{84} & 84  & \2{97} & 87   & 92 &  83 &  \2{91} & \2{81} &  94 & \2{85} & \2{88} &  \2{93} &   \2{90} &  \2{95} & \2{85}  \\ 
{\bf CRL(I)* } & \2{95} &\2{95}  &\1{84}  &\2{74}  &\1{72} &\1{90} &\1{87} &\2{88}  &96  & 88 &\1{96} &\1{87} &\1{92} &\1{85} &\2{98} &\1{89}  &\1{92} &\1{95} &\1{94} &\1{97}  &\bf \1{86}\\
\hline
		
	\end{tabular}
	\vspace{-0.5 cm}
\end{table*}


\noindent {\bf Datasets \& Performance Metric.}
As shown in Table~\ref{tab:dataset_imbalance}, both CelebA and X-Domain datasets are highly
imbalanced. For that reason, we selected these two datasets for our
evaluations. The CelebA~\cite{liu2015deep} facial attribute
dataset has 202,599 web images
from 10,177 person identities with per person on average 20 images. Each face image
is annotated by 40 binary attributes. 
The X-Domain \cite{chen2015deep} clothing attribute
dataset\footnote{We did not select the DeepFashion
  \cite{liu2016deepfashion} dataset for our evaluation because this dataset is
  relatively well balanced compared to X-Domain (Table~\ref{tab:dataset_imbalance}), due to the strict
  data cleaning process applied.} consists of 245,467 shop images from online retailers like {\it Tmall.com}.
Each clothing image is annotated by $\leq$ $9$ attribute categories
and each category has a different set of values (mutually
exclusive within each set)
ranging from $6$ (slv-len) to $55$ (colour).
In total, there are $178$ distinctive attribute values %
{in 9 categories (labels)}.
For each attribute label, we adopted the {\em class-imbalanced} accuracy 
(i.e. mean sensitivity) 
as the model performance metric given imbalanced data \cite{fernandez2011dynamic,huang2016learning}.
This additionally considers the class distribution statistics
in performance measurement. 
%

\subsection{Evaluation on Imbalanced Face Attributes}
\label{sec:eval_face_attributes}
\noindent {\bf Competitors.} 
We compared CRL against 8 existing methods including 4 state-of-the-art
deep models for facial attribute recognition on CelebA:
(1) Over-Sampling \cite{drummond2003c4}, (2) Down-Sampling
\cite{drummond2003c4}, (3) Cost-Sensitive \cite{he2009learning}, (4)
Large Margin Local Embedding (LMLE) \cite{huang2016learning}, (5)
PANDA \cite{zhang2014panda}, (6) ANet \cite{liu2015deep}, (7)
Triplet-$k$NN \cite{schroff2015facenet}, and (8) DeepID2
\cite{sun2014deep}.

\noindent {\bf Training/Test Data Partition.}	
We adopted the same data partition on CelebA as in \cite{liu2015deep,huang2016learning}:
The first 162,770 images are used for training (10,000 images for
validation), the following 19,867 images for training the SVM
classifiers required by PANDA \cite{zhang2014panda} and ANet
\cite{liu2015deep} models, and the remaining 19,962 images for testing. Note that identities of
all face images are non-overlapped in this partition.

\noindent {\bf Network Architecture \& Parameter Settings.}
We adopted the five layers CNN network architecture of DeepID2 \cite{sun2014deep} as
the basis for training all six imbalanced data learning methods including
both our CRL models (C\&I), the same for LMLE as reported in~\cite{huang2016learning}.
In addition to the DeepID2's shared FC$_1$ layer, for explicitly modelling the attribute
specificness, in our CRL model we added a respective 64-dimensional FC$_2$ layer 
for each face attribute, in the spirit of multi-task learning
\cite{evgeniou2004regularized,ando2005framework}. 
We set the learning rate at $0.001$
to train our model from scratch on the CelebA
face images.
We fixed the decay to $0.0005$
and the momentum to $0.9$. 
Our CRL model converges after 200 epochs training with a batchsize of 128
images.

\begin{figure}[th] 
	\centering
	\includegraphics[width=1\linewidth]{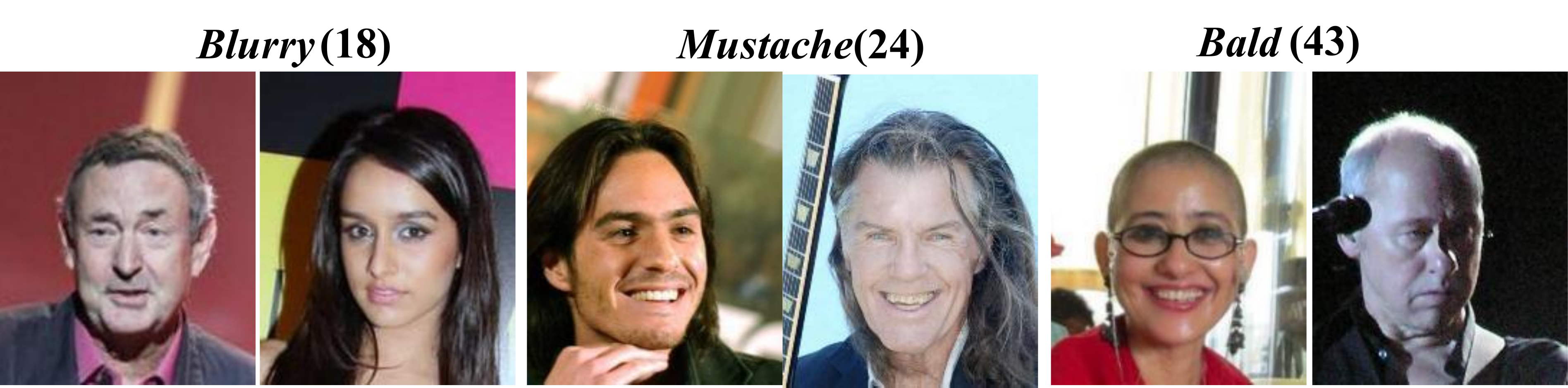}
	\vskip -0.1cm
	\caption{
		Examples (3 pairs) of facial attribute
                recognition (imbalance ratio in bracket). 
		In each pair, DeepID2 missed both, whilst CRL
                identified the left image but failed the right
                image.
	}
	\label{fig:vis_face}
	\vspace{-0.4 cm}
\end{figure}

\noindent {\bf Comparative Evaluation.} 
Facial attribute recognition performance comparisons are shown
in Table \ref{tab:arts_face}.
It is evident that CRL outperforms on average accuracy
all competitors including the state-of-the-art
attribute recognition models and imbalanced data learning methods 
Compared to the best non-imbalanced learning model DeepID2, CRL(I)
improves average accuracy by $5\%$. Compared to the state-of-the-art imbalanced
learning model LMLE, CRL(I) is better by $2\%$ in average accuracy. Other classical
imbalanced learning methods perform similarly to DeepID2.
The performance drop by Down-Sampling is due to discarding
useful data for balancing distributions.
%
This demonstrates the importance of explicit imbalanced data learning,
and the superiority of the proposed batch incremental class
rectification hard mining approach to handling imbalanced data over alternative methods. 
Figure~\ref{fig:vis_face} shows qualitative examples.
%

\begin{figure} [h]
	\centering
	\includegraphics[width=1\linewidth]{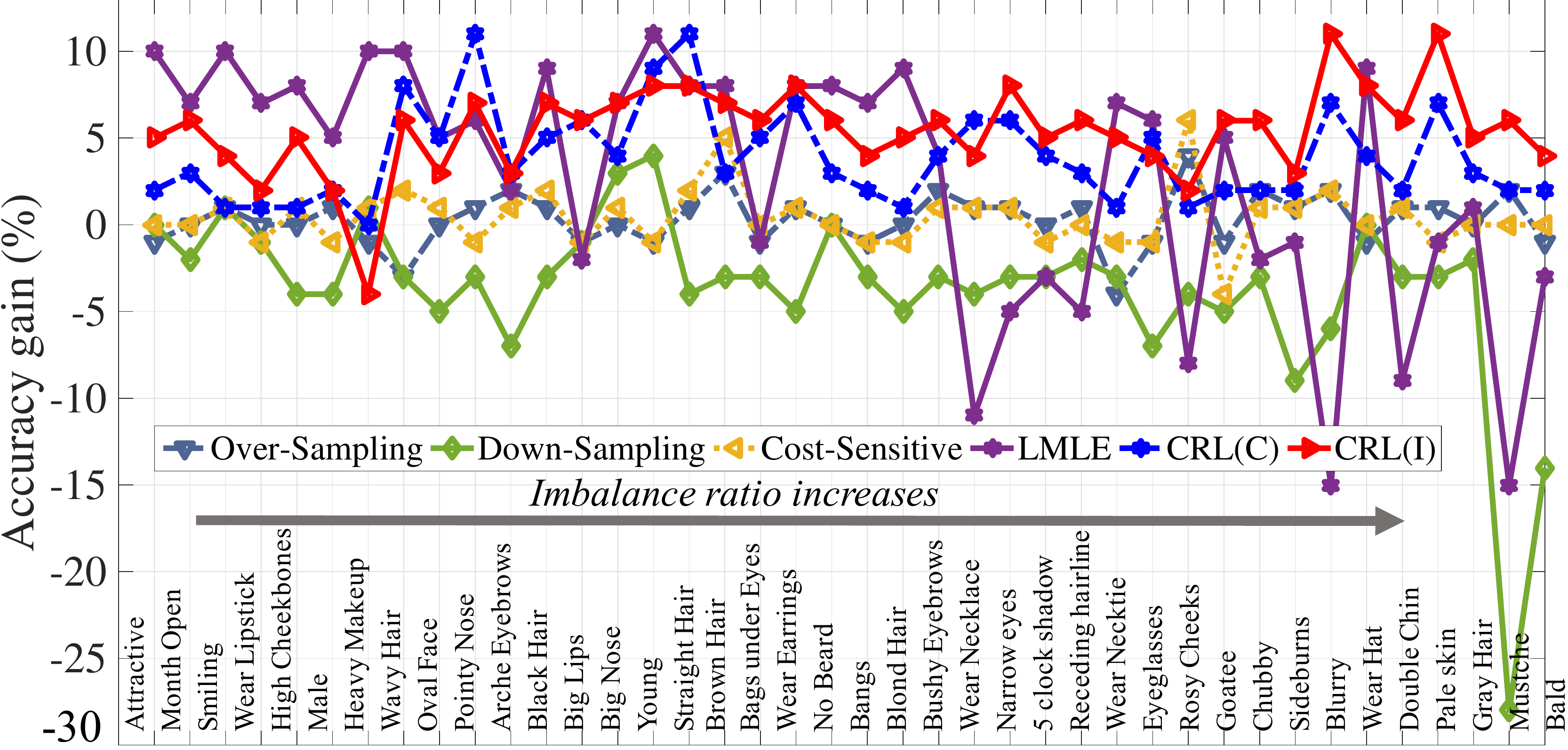}
	\vskip -0.1cm
	\caption{
		{{Performance gain over DeepID2 \cite{sun2014deep} by the
				six imbalanced learning methods on
                                the 40 CelebA facial attributes \cite{liu2015deep}.
				Attributes sorted from left to right
                                in increasing imbalance ratio.}	
		}
		}
	\label{fig:accgain_face}
	\vspace{-0.1cm}
\end{figure}

\noindent {\bf Model Performance vs. Data Imbalance Ratio.}
Figure~\ref{fig:accgain_face} 
further shows the accuracy gain of six imbalanced learning
methods. It can be seen that LMLE copes
better with less imbalanced attributes (towards the left side in
Figure~\ref{fig:accgain_face}), but degrades notably given higher data imbalance ratio.
Also, LMLE performs worse than DeepID2
on more attributes towards the right of ``Wear Necklace'' in
Figure~\ref{fig:accgain_face}, i.e. imbalance ratio greater than 1:7
in Table~\ref{tab:arts_face}. 
In contrast, CRLs with both class-level (CRL(C)) and instance-level
(CRL(I)) hard mining perform particularly well on attributes with high
imbalance ratios. More importantly, even though CRL(I) only outperforms LMLE by 2\% in average accuracy 
over all 40 attributes, this margin increases to 7\% in
average accuracy over the {\em 20 most} imbalanced attributes. Moreover, on some of
the very imbalanced attributes, CRL(I) outperforms LMLE by 21\% on
``Mustache" and 26\% on ``Blurry". Interestingly, the ``Blurry'' attribute is
challenging due to its global characteristics therefore not defined by
local features and very subtle, similar to the ``Mustache'' attribute
(see Figure~\ref{fig:vis_face}).
This demonstrates that CRL is
significantly better than LMLE in coping with severely imbalanced data
learning. This becomes more evident with the X-domain
clothing attributes (Sec.~\ref{sec:eval_clothing_attr}), mainly
because given severe imbalanced data,
it is difficult for LMLE to cluster effectively due to very few minority
class samples, which leads to inaccurate classification feature learning. 
%

\noindent {\bf Model Training Time.}
We also tested the training time cost of LMLE independently on an identical hardware setup as for CRL: LMLE took 388 hours to train whilst CRL (C/I) took 27/35 hours respectively with 11 times training costs advantage over LMLE in practice. Specifically, LMLE needs 4 rounds of quintuplets construction with each taking 96 hours, and 4 rounds of deep model learning with each taking 1 hour. In total, 4 * (96+1) = 388 hours.


\begin{table*} [h] 
	\footnotesize
	\centering
	\setlength{\tabcolsep}{0.2cm}
	\caption{Clothing attributes recognition on the X-Domain dataset. 
		* Imbalanced data learning models. 
		Metric: Class-balanced accuracy, i.e. mean sensitivity (\%)..
		CRL(C/I): CRL with Class/Instance level hard mining.
		Slv-Shp: Sleeve-Shape; Slv-Len: Sleeve-Length.
		The $1^\text{st}/2^\text{nd}$ best results are highlighted in red/blue.}
	\label{tab:arts_clothing}
	\begin{tabular}{c||c|c|c|c|c|c|c|c|c||c}
		\hline
		\backslashbox{\bf Methods}{\bf Attributes}
	 
	& Category & Colour & Collar & Button  & Pattern & Shape & Length  & Slv-Shp& Slv-Len & \bf {Average}\\\hline
	\bf Imbalance ratio (1:x)\footnote{All attributes are grouped into 9 branches and inner group they are mutually exclusive, and by doing this, imbalance degree inner branch may decrease (less than 1:10289). }&2&138&210&242&476&2138&3401&4115&4162 & \\\hline
		DDAN \cite{chen2015deep} 
		&46.12 &31.28&22.44&40.21    &29.54 &23.21&32.22  &19.53&40.21   & {\bf 31.64}\\
		FashionNet \cite{liu2016deepfashion} 
		&48.45 &36.82&25.27&43.85     &31.60 &27.37&38.56&20.53 &45.16   & {\bf 35.29}\\
		
		DARN \cite{huang2015cross} 
		&65.63  &44.20&31.79&58.30 &44.98  & 28.57 &45.10 & 18.88&51.74    &{\bf 43.24}\\
		
		MTCT \cite{dong2016multi} 
		&72.51 &74.68&70.54&76.28&76.34&68.84&77.89& 67.45&77.21  &{\bf 73.53}\\ \hline

		Over-Sampling* \cite{jeatrakul2010classification}
		&73.34 &75.12&71.66&77.35&77.52&68.98&78.66& 67.90&78.19& {\bf 74.30}  \\  
		
		Down-Sampling* \cite{mani2003knn}
		&49.21&33.19&19.67&33.11&22.22&30.33&23.27&12.49&13.10&{\bf 26.29}\\
		
		Cost-Sensitive* \cite{he2009learning} 
		&76.07 &77.71&71.24&79.19&77.37&69.08&78.08& 67.53&77.17&{\bf 74.49} \\ 
		
		LMLE* \cite{huang2016learning}&75.90&77.62&70.84&78.67&77.83&71.27&79.14&69.83&80.83&{\bf 75.77} \\
		
		{\bf CRL(C)}* 
		& \2{76.85}	&\2{79.61}	&\2{74.40}	&\2{81.01}	&\2{81.19}	&\2{73.36} &\2{81.71}	&\2{74.06}	&\2{81.99}&\2{\bf 78.24}\\
		{\bf CRL(I)}*  
		&\1{77.41} &\1{81.50} & \1{76.60}&\1{81.10} &\1{82.31} &\1{74.56} &\1{83.05} &\1{75.49} &\1{84.92} &\1{\bf 79.66} \\
		\hline
	\end{tabular}
	\vspace{-0.6cm}
	
\end{table*}

\subsection{Evaluation on Imbalanced Clothing Attributes}
\label{sec:eval_clothing_attr}

\noindent {\bf Competitors.}
In addition to the four imbalanced learning methods
(Over-Sampling, Down-Sampling, Cost-Sensitive, LMLE\footnote{We trained an independent LMLE CNN model for each attribute label. This is because the quintuplets construction over all attribute labels is prohibitively expensive in terms of computing cost.}) used for face
attribute evaluation,
we also compared against four other state-of-the-arts clothing attribute recognition
models: 
(1) Deep Domain Adaptation Network (DDAN) \cite{chen2015deep}, (2)
{Dual Attribute-aware Ranking Network (DARN) \cite{huang2015cross},
  (3) FashionNet \cite{liu2016deepfashion}, and (4) Multi-Task Curriculum
  Transfer (MTCT) \cite{dong2016multi}.

\noindent {\bf Training/Test Data Partition.}
We adopted the same data partition as in \cite{huang2015cross,dong2016multi}: 
Randomly selecting 165,467 clothing images 
for training
and the remaining 80,000 for testing. 

{\noindent {\bf Network Architecture.}
	We used the same network structure as the MTCT \cite{dong2016multi}. 
	Specifically, this network is composited of 
	five stacked NIN conv units \cite{lin2013network} and 
	$n_\text{attr}$ parallel branches with each 
	a three FC layers sub-network for modelling
	one of the $n_\text{attr}$ attributes respectively, 
	in the spirit of multi-task learning
	\cite{evgeniou2004regularized,ando2005framework}.

\noindent {\bf Parameter Settings.}
We pre-trained a base model 
on ImageNet-1K at
the learning rate $0.001$,
and then finetuned the CRL model on the X-Domain
clothing images at the same rate $0.001$.
We fixed the decay to $0.0005$
and the momentum to $0.9$.
The CRL model converges after 150 epochs.  
The batchsize is 256.

%
\noindent {\bf Comparative Evaluation.}
Table \ref{tab:arts_clothing} shows the comparative evaluation of 10 different
models on the X-Domain benchmark dataset.
It is evident that CRL(I) surpasses all
other models on all attribute categories.
This shows the significant superiority and scalability of the class
rectification hard mining with batch incremental approach in coping
with extremely imbalanced attribute data, with the maximal imbalance
ratio $4,162$ vs. $43$ in CelebA attributes (Figure~\ref{fig:problem}).
A lack of explicit imbalanced learning mechanism in other models such
as DDAN, FashionNet, DARN and MTCT suffers notably. 
Among the 6 models designed for imbalance data learning, 
we can observe similar trends as in face attribute recognition on
CelebA. Whilst LMLE improves notably on classic imbalanced data
learning methods, it remains inferior to CRL(I) by significant
margins (4\% in accuracy over all attributes).

\begin{figure}[th] 
	\includegraphics[width=1\linewidth]{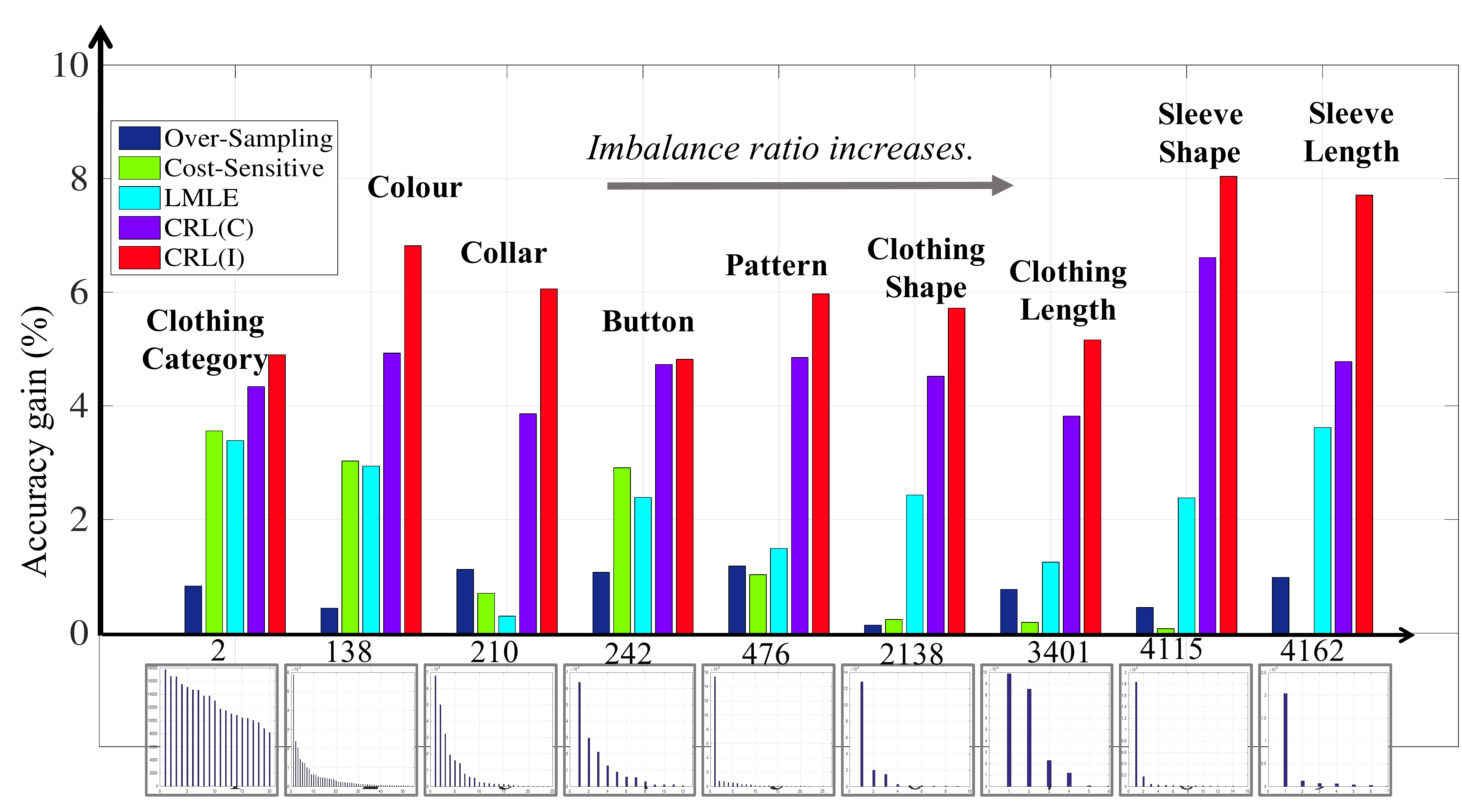}
	 \vskip -0.1cm
	\caption{
		Model performance {\em additional} gain over the MTCT on 9 clothing attributes with increasing
		imbalance ratios on X-Domain. 
		}
	\label{fig:accgain}
	\vspace{-0.4 cm}
\end{figure}

\noindent {\bf Model Effectiveness in Mitigating Data Imbalance.}
We compared the relative
performance gain of the 6 different imbalanced data learning models
(Down-Sampling was excluded due to poor performance) 
against the MTCT (as the baseline), 
along with 
the imbalance ratio for each clothing attribute.
Figure \ref{fig:accgain} shows the comparisons and it is
evident that CRL is clearly superior in learning
severely imbalanced attributes, e.g. on ``Sleeve
Shape'', CRL(C) and CRL(I) achieve $8\%$ and $7\%$ accuracy gain over MTCT respectively, as compared
to the second best LMLE obtaining only $2\%$
improvement. Qualitative examples are shown in Figure~\ref{fig:vis_cloth}.
\vspace{-0.4 cm}
\begin{figure}[th] 
	\centering
	\includegraphics[width=0.95\linewidth]{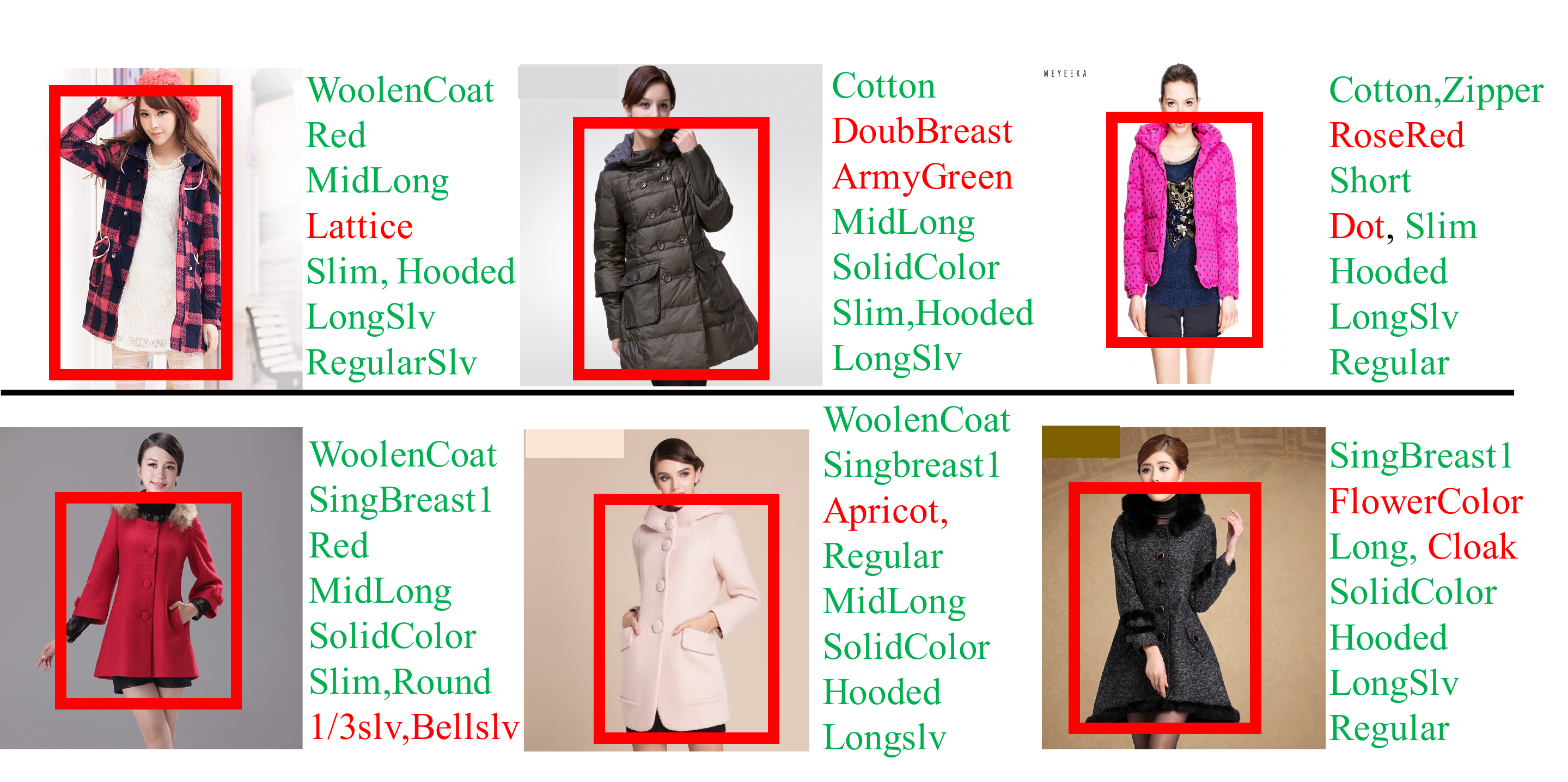}
	 \vskip -0.2cm
	\caption{
		Examples of clothing attribute recognition by the CRL(I) model,
		with falsely predicted attributes in red.
	}
	\label{fig:vis_cloth}
	\vspace{-0.5cm}
\end{figure}

\subsection{Analysis on Rectification Loss and Hard Mining}
\label{sec:eval_CRL_loss}
We evaluated the effects of two different hard mining schemes (Class
and Instance level) (Sec.~\ref{sec:method_hard_mining}), and 
three different CRL loss functions (Relative, Absolute, and
Distribution comparisons) (Sec. \ref{sec:method_CRL_function}).
In total, we tested 6 different CRL variants.
We evaluated the performance of these 6 CRL models
by the accuracy gain over a non-imbalance learning baseline model: DeepID2 on CelebA and MTCT on X-domain. 
It is evident from Table \ref{tab:crossdata} that:
(1) All CRL models improve accuracy on both facial and clothing
attribute recognition.
(2) For both face and clothing, CRL(I+R) is the best and its
performance advantage over other models is doubled on the more
imbalanced X-Domain when compared to that on CelebA.
(3) Most CRL models achieve greater performance gains on X-Domain than on CelebA. 
(4) Using the same loss function, instance-level hard mining is superior in most cases.

\begin{table}[th]
\centering
\footnotesize
\caption{Comparing different hard mining schemes (Class and Instance level) and 
	loss functions (Relative({\bf R}), Absolute({\bf A}), and Distribution({\bf D})). 
Metric: additional gain in average accuracy (\%).}
\label{tab:crossdata}
\begin{tabular}{c|c|c|c|c|c|c}
\hline
\multirow{1}{*}{Dataset} & \multicolumn{3}{c|}{CelebA} & \multicolumn{3}{c}{X-domain} \\\hline \cline{1-7} 
Loss function & A & R & D & A &R & D \\ \hline \hline
Class Level & 5.71 & 4.23 & 0.54 & 3.46 & 4.71 & 1.20 \\ 
Instance Level & 5.67 & \textbf{5.85} & 2.12 & 4.92 & \textbf{6.13} & 2.05 \\ \hline
\end{tabular}
\end{table}
\vspace{-0.5 cm}
\section{Conclusion}
In this work, we formulated an end-to-end imbalanced deep learning framework for 
clothing and facial attribute recognition with very large scale imbalanced
training data. The proposed Class Rectification Loss (CRL) model with batch-wise
incremental hard positive and negative mining of the minority classes
is designed to regularise deep model learning behaviour given training
data with significantly imbalanced class distributions in very large
scale data. 
%
%
%
Our experiments show clear advantages of the proposed CRL model over not only the state-of-the-art imbalanced data learning models but also dedicated attribute recognition methods for
multi-label clothing and facial attribute recognition, surpassing the
state-of-the-art LMLE model by $2\%$ in average accuracy on the CelebA face benchmark and 
$4\%$ on the more imbalanced X-Domain clothing benchmark, whilst having over
threes time faster model training time advantage.

\section*{Acknowledgements}
\vspace{-0.2cm}
\noindent This work was partially supported by the China Scholarship Council, Vision Semantics Ltd., and the Royal Society Newton Advanced Fellowship Programme (NA150459). 

{\small
\bibliographystyle{ieee}
\bibliography{clothingAttr}
}

\end{document}